\documentclass{article}

\usepackage{arxiv}

\usepackage[utf8]{inputenc} 
\usepackage[T1]{fontenc}    
\usepackage{hyperref}       
\usepackage{url}            
\usepackage{booktabs}       
\usepackage{amsfonts}       
\usepackage{nicefrac}       
\usepackage{microtype}      
\usepackage{lipsum}		
\usepackage{graphicx}
\usepackage{natbib}
\usepackage{doi}
\DeclareFixedFont{\myfontb}{OT1}{ptm}{bx}{n}{10pt}
\usepackage{threeparttable}
\usepackage{multirow}
\usepackage{color}
\usepackage{authblk}

\title{Adaptively Customizing Activation Functions for Various Layers}

\date{}

\renewcommand{\shorttitle}{\textit{arXiv} Adaptively Customizing Activation Functions for Various Layers}

\hypersetup{
pdftitle={Adaptively Customizing Activation Functions for Various Layers},
pdfauthor={Haigen~Hu, Aizhu~Liu},
pdfkeywords={Adaptive activation function, Adaptable parameters, Various layers, Deep learning},
}

\author[1,2]{Haigen Hu} 
\author[1,2]{Aizhu Liu}
\author[1,2]{Qiu Guan}
\author[1,2]{Xiaoxin Li}
\author[1,3]{Shengyong Chen} 
\author[1,2,*]{Qianwei Zhou}
\affil[1]{College of Computer Science and Technology, Zhejiang University of Technology}
\affil[2]{Key Laboratory of Visual Media Intelligent Processing Technology of Zhejiang Province}
\affil[3]{School of Computer Science and Engineering, Tianjin University of Technology}

\begin{document}

\maketitle

\begin{abstract}
To enhance the nonlinearity of neural networks and increase their mapping abilities between the inputs and response variables, activation functions play a crucial role to model more complex relationships and patterns in the data. In this work, a novel methodology is proposed to adaptively customize activation functions only by adding very few parameters to the traditional activation functions such as Sigmoid, Tanh, and ReLU. To verify the effectiveness of the proposed methodology, some theoretical and experimental analysis on accelerating the convergence and improving the performance is presented, and a series of experiments are conducted based on various network models (such as AlexNet, VGGNet, GoogLeNet, ResNet and DenseNet),  and various datasets (such as CIFAR10, CIFAR100, miniImageNet, PASCAL VOC and COCO) . To further verify the validity and suitability in various optimization strategies and usage scenarios, some comparison experiments are also implemented among different optimization strategies (such as SGD, Momentum, AdaGrad, AdaDelta and ADAM) and different recognition tasks like classification and detection. The results show that the proposed methodology is very simple but with significant performance in convergence speed, precision and generalization, and it can surpass other popular methods like ReLU and adaptive functions like Swish in almost all experiments in terms of overall performance.The code is publicly available at \href{https://github.com/HuHaigen/Adaptively-Customizing-Activation-Functions}{https://github.com/HuHaigen/Adaptively-Customizing-Activation-Functions}. The package includes the proposed three adaptive activation functions for reproducibility purposes.
\end{abstract}


\section{Introduction}
Activation functions play a key role during the training process of neural networks, and considerable attention has been paid to explore standard activation functions over the past years. Especially, with the remarkable development of Deep Neural Networks (DNN) in various computer vision applications, such as image classification \citep{he2016deep,krizhevsky2012imagenet,tan2017photograph}, image segmentation \citep{chen2017deeplab}, object detection \citep{girshick2014rich,jiang2016speed,he2015delving}, image enhancement \citep{lin2018image,tang2018joint}, image retrieval \citep{yu2014click,yu2016deep} and tracking \citep{wu2016regional}, Rectified Linear Unit (ReLU) \citep{Nair2010} has become extremely popular in the deep learning community in recent years. Owing to the significant improvements of ReLU in the deep neural networks, some extended versions are constantly springing up. For instance, Leaky ReLU (LReLU) \citep{maas2013rectifier} is proposed by replacing the negative part of the ReLU with a non-zero slope, while Exponential Linear Units (ELUs) \citep{clevert2015fast} can tend to converge cost to zero faster and produce more accurate results. All these extended versions can more or less achieve a certain effect in the respective fields. 

However, there is hardly a generally accepted rule-of-thumb for the choice of activation functions owing to the fact that it solely depends on the problem at hand. Even the most popularly and commonly used activation function ReLU is not suitable for all datasets and network architectures. Therefore, adaptive activation functions have drawn more and more attention in recent years. For example, Maxout\citep{goodfellow2013maxout} can approximate any convex functions by selecting the maximum output value of multiple linear activation functions, but a large number of extra parameters are introduced, which causes large storage memory and high computation cost. In Parametric rectified linear unit (PReLU) \citep{he2015delving}, the slopes of negative part can be obtained by learning from data rather than the pre-defined fixed values, thus PReLU has theoretically all the advantages of ReLU and effectively avoids Dead ReLU. But in practice, it has not been fully confirmed that PReLU always surpasses ReLU. In 2017, an activation function with the property of  ``self-normalization" is proposed, named SELU \citep{klambauer2017self}, and it can avoid the problem of gradient vanishing and exploding, thereby leading to the feedforward neural network to obtain beyond state-of-the-art performance. However, the effectiveness of SELU in Convolutional neural networks (CNN) has not been confirmed. In the same year, swish \citep{ramachandran2017searching} with some complex characteristics, such as no upper and lower bound, smooth and non-monotonic, can perform better than ReLU on many deep models.

Although the existing adaptive activation functions are relatively more flexible than the traditional activation function owing to the adaptability, and have already achieved great improvements, they are limited to some specific application scenarios, and there are still many problems to be solved, such as low generalization capability and poor precision performance. For example, their performance often depends on some specific network models and data sets. In this work, a novel methodology is proposed to explore the optimal activation functions with more flexibility and adaptability only by adding few additional parameters to the traditional activation functions such as Sigmoid, Tanh and ReLU. The proposed methodology can avoid local minimums and accelerate convergence only by introducing very few parameters to the fixed activation functions, thereby increasing the precision, reducing the training cost and improving the generalization performance.

The primary contributions of our work are summarized as follows: 
\begin{itemize}
	\item A novel methodology is proposed to customize activation functions with more flexibility and adaptation for various layers only by introducing very few parameters to the traditional activation functions such as Sigmoid, Tanh, and ReLU. 
	\item A theoretical analysis for accelerating the convergence and improving the performance is presented by taking an activation function of one layer as an example without loss of generality, and an experimental study is performed by comparing the weight increments between two successive epochs in different layers during the training process between the proposed AReLU and ReLU on CIFAR100 based on VGGNet.
	\item The proposed \textit{AReLU} is a generalized form of the \textit{ReLU-based} versions, while \textit{ReLU} and \textit{PReLU} are the special cases of the proposed \textit{AReLU}.
\end{itemize}

The rest of the paper is organised as follows. Section 2 introduces the related work, and the proposed methodology is presented in Section 3. Section 4 presents the analysis for our methodology. Section 5 details the experimental results for comparison and validation. Section 6 concludes the paper.

\section{Related work}

Over the last few decades, many various activation functions have been proposed in the artificial neural network community. According to whether the parameter or shape of an activation function is learnable or variable during the training phase, activation functions can be divided in two categories: fixed activation functions and adaptive activation functions.


\subsection{Fixed activation functions}

\begin{figure*}[!t]
	\centering
	\includegraphics[width=1.6 in]{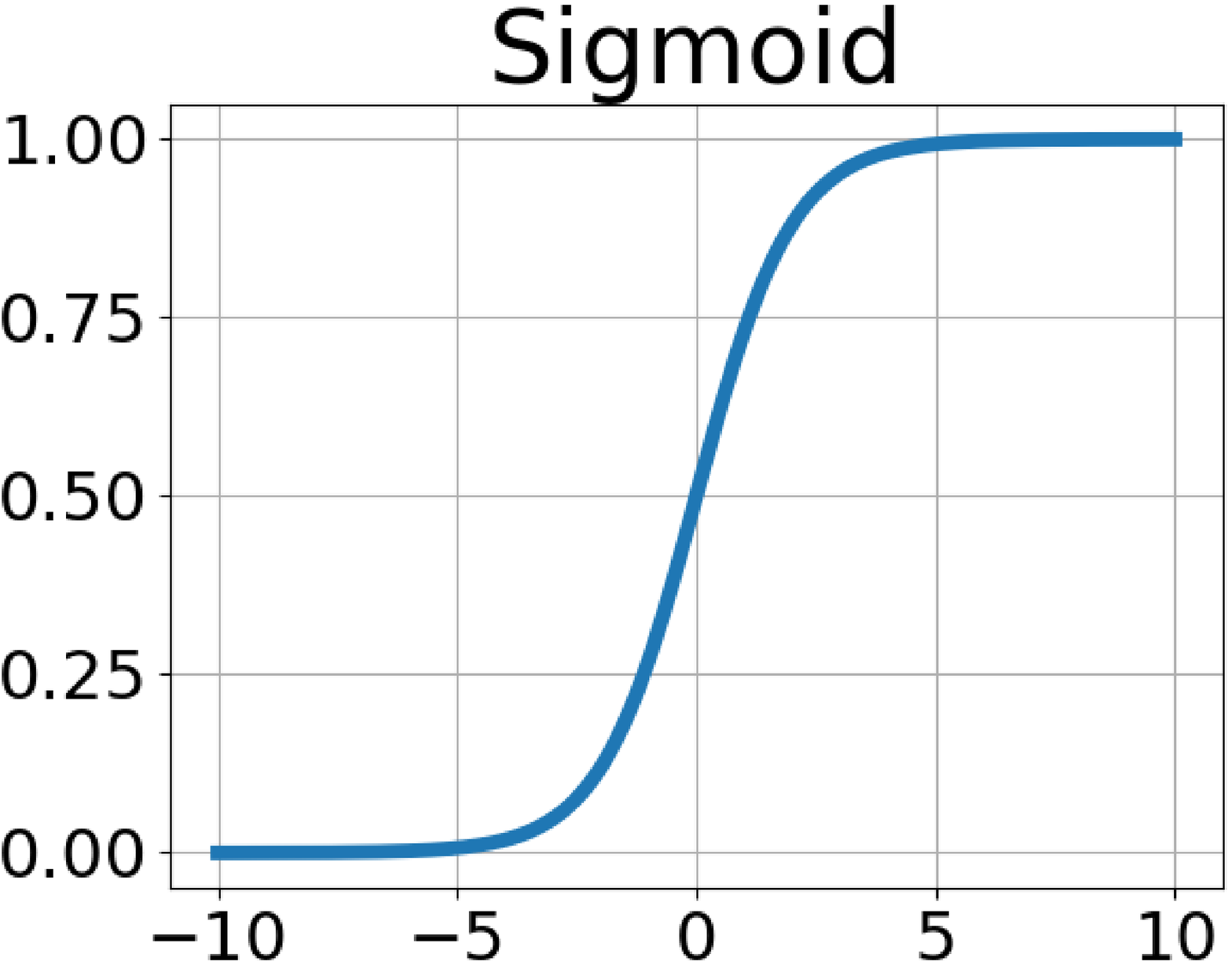}
    \includegraphics[width=1.6 in]{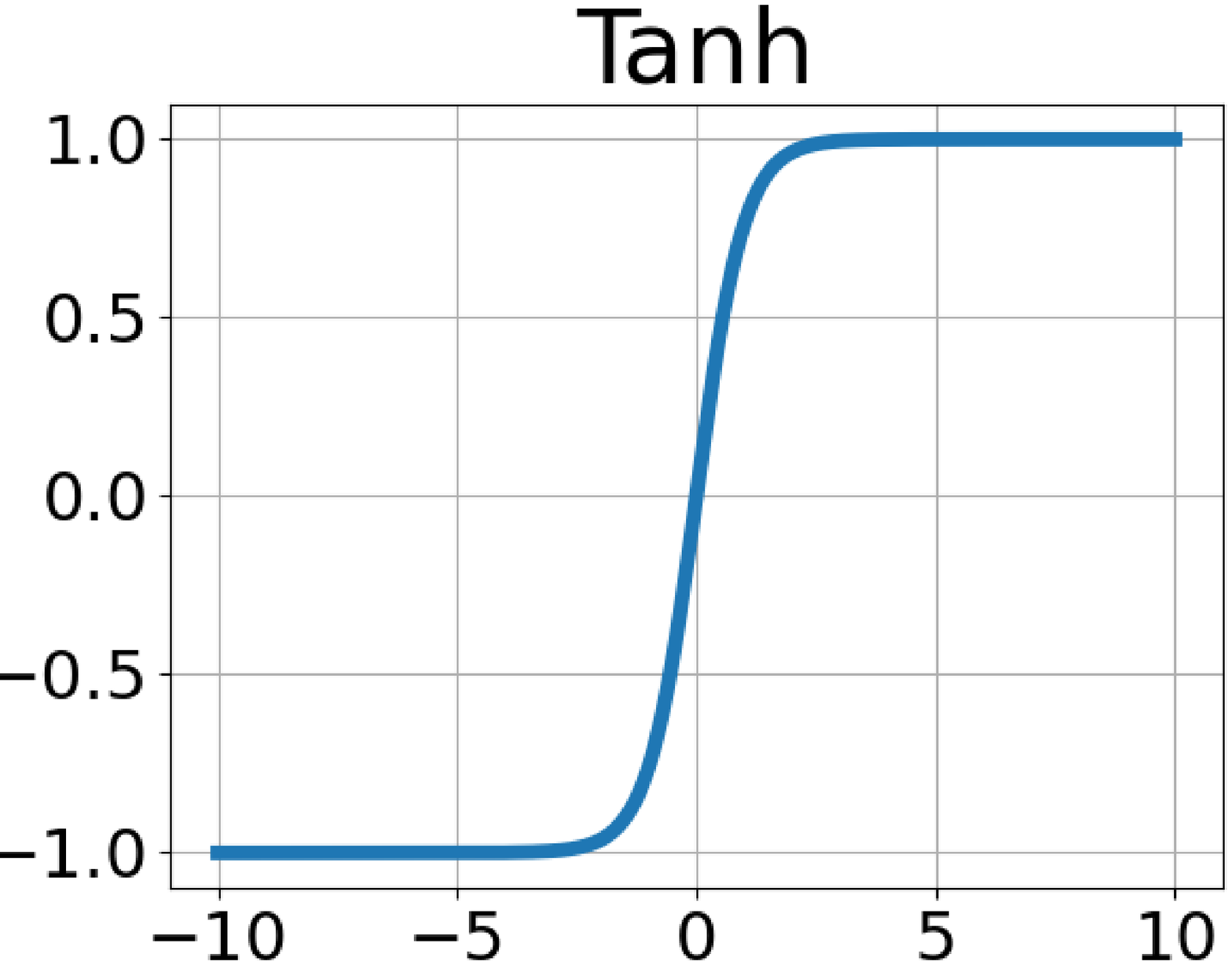}
    \includegraphics[width=1.6 in]{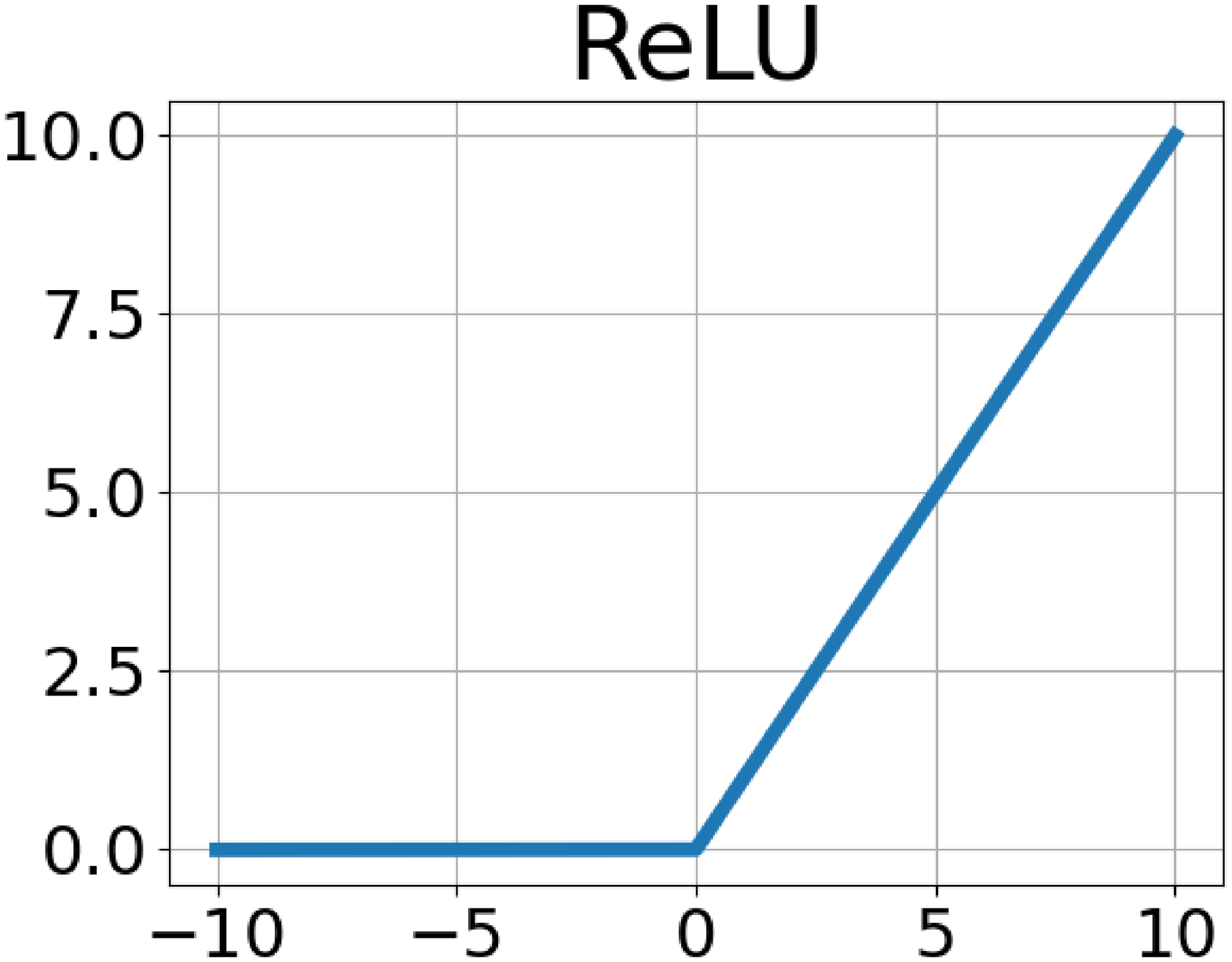}
	\caption{An illustration of fixed activation functions with a fixed shape.}
	\label{fig:fig1}
\end{figure*}

Fixed activation functions indicate that the parameters or shapes can not be modified during the training phase (shown in Fig. \ref{fig:fig1}), and the most common fixed activation functions can be fallen into three categories: Logistic (Sigmoid), Hyperbolic Tangent (Tanh) and Rectified Linear Activation (ReLU).

\paragraph{Sigmoid}
Sigmoid function is a common S-like function or S-like growth curve, and is normally used to refer specifically to the logistic function. It can map any real value to the range [0,1], thereby being interpreted as a probability, defined as follows: 
\begin{equation}
Sigmoid(x)=\sigma(x)=\frac{1}{1+e^{-x}}
\end{equation}

It is differentiable, and the derivative is derived as follows:
\begin{equation}
\sigma'(x)=\sigma(x)(1-\sigma(x))
\end{equation}

Note that the gradient $\sigma'(x) \to 0$ as $\sigma(x) \to 0$ or $\sigma(x) \to 1$, meaning that, when the output of Sigmoid saturates for a large positive or negative inputs (i.e. the curve becomes parallel to $x$-axis shown in Fig. \ref{fig:fig1}), the gradients are almost zero. Due to the zero gradient, the weights are no longer updated and the networks will not learn, thus the neuron dies, thereby causing the vanishing gradient problem. Besides, Sigmoid outputs are not zero-centered, and it can indirectly introduce undesirable zig-zagging dynamics in the gradient updates for the weights.

\paragraph{Tanh}
Tanh function, a hyperbolic tangent function, graphically looks very similar to Sigmoid. Actually, the Tanh is simply a scaled Sigmoid, such that its outputs range from -1 to 1, defined as follows:
\begin{equation}
Tanh(x)=\frac{sinh(x)}{cosh(x)}=\frac{e^x-e^{-x}}{e^x+e^{-x}}=2\sigma(2x)- 1
\end{equation}

Like the Sigmoid, Tanh is also affected by the vanishing gradient problem. But unlike the Sigmoid, its output is zero-centered,  the negative inputs will be mapped strongly negatives and the zero inputs will be mapped near zero. Therefore, the non-linearity of Tanh is always preferred to that of Sigmoid, and it has been widely used in deep learning \& machine learning, especially in classification scenarios between two classes.

\paragraph{ReLU}
ReLU is a very simple and efficient activation function that has been widely used in almost all deep learning domains, especially in CNNs, defined as 
\begin{equation}
ReLU(x)=max(0,x)
\end{equation}

Owing to the simpler mathematical operations, ReLU is far more computationally efficient than Tanh and Sigmoid. Besides, ReLU can solve parts of the saturation problem only in the positive region. Whereas for the negative inputs, the results contain one or more true zero values (called a sparse representation) to accelerate learning and simplify the model in representational learning, but the weights and biases are not updated owing to the zero gradient during the backpropogation process, thereby causing the dying ReLU problem. 

\subsection{Adaptive activation functions}
\begin{figure*}[!t]
	\centering
	\includegraphics[width=1.5 in]{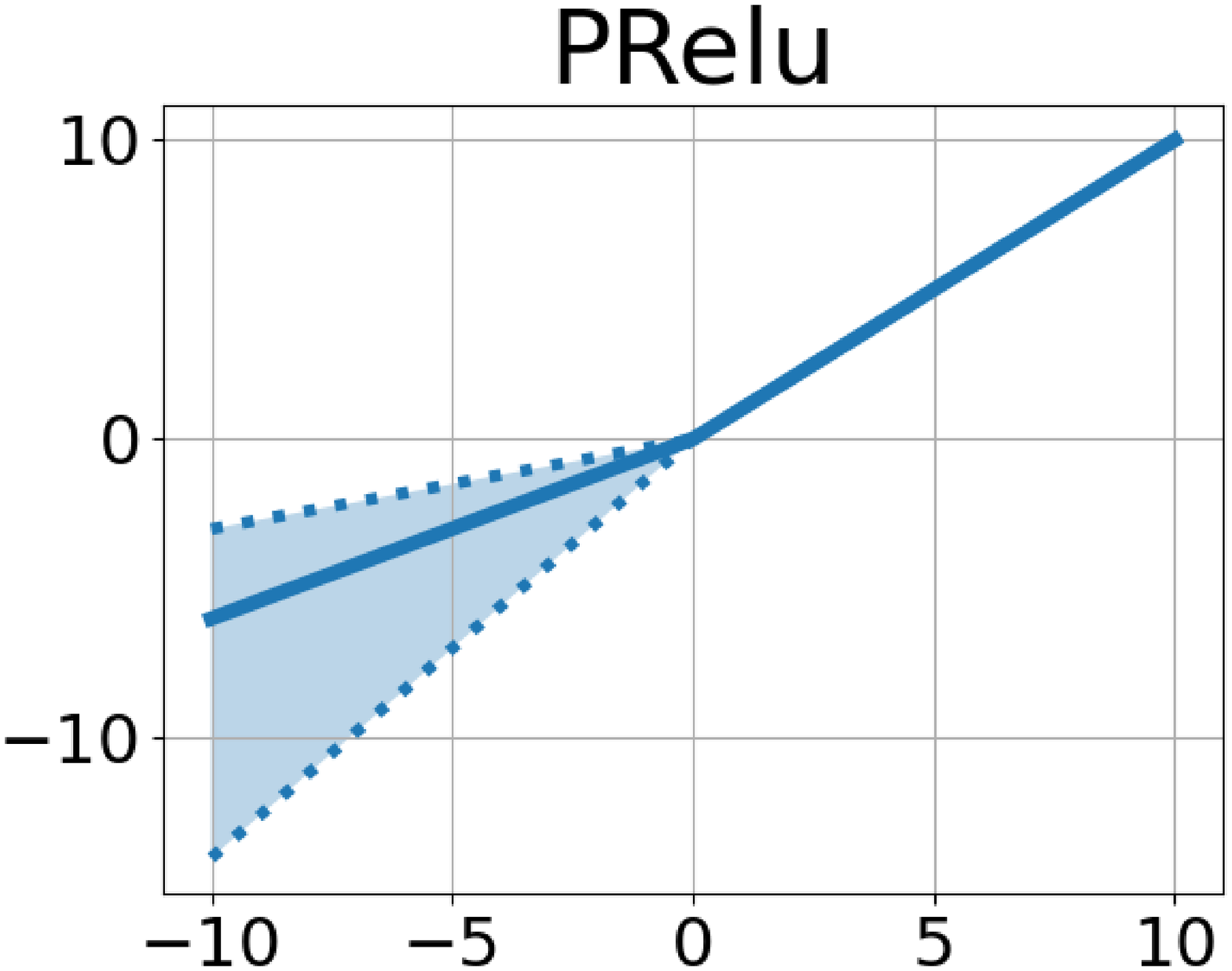}
    \includegraphics[width=1.5 in]{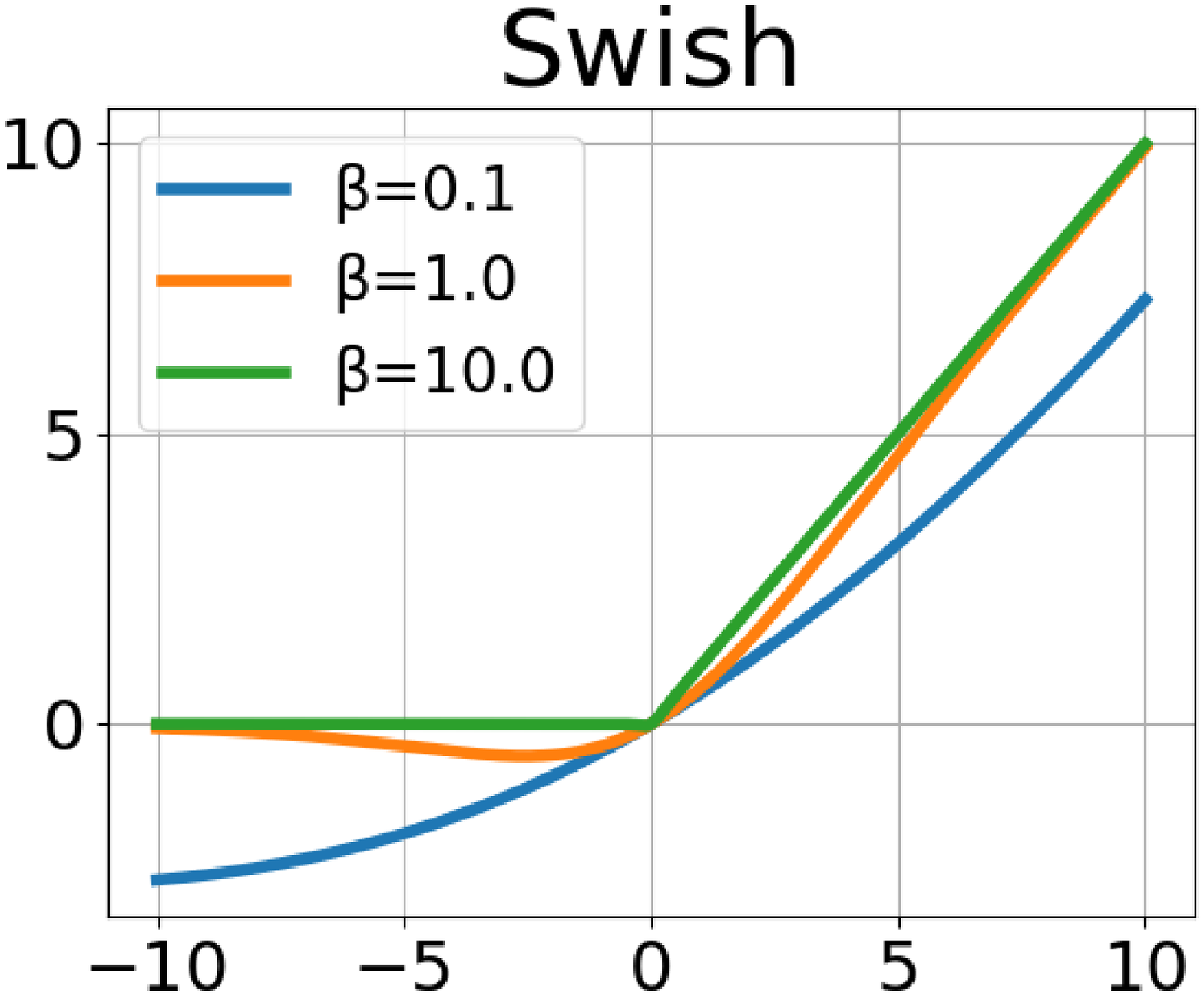}
    \includegraphics[width=1.5 in]{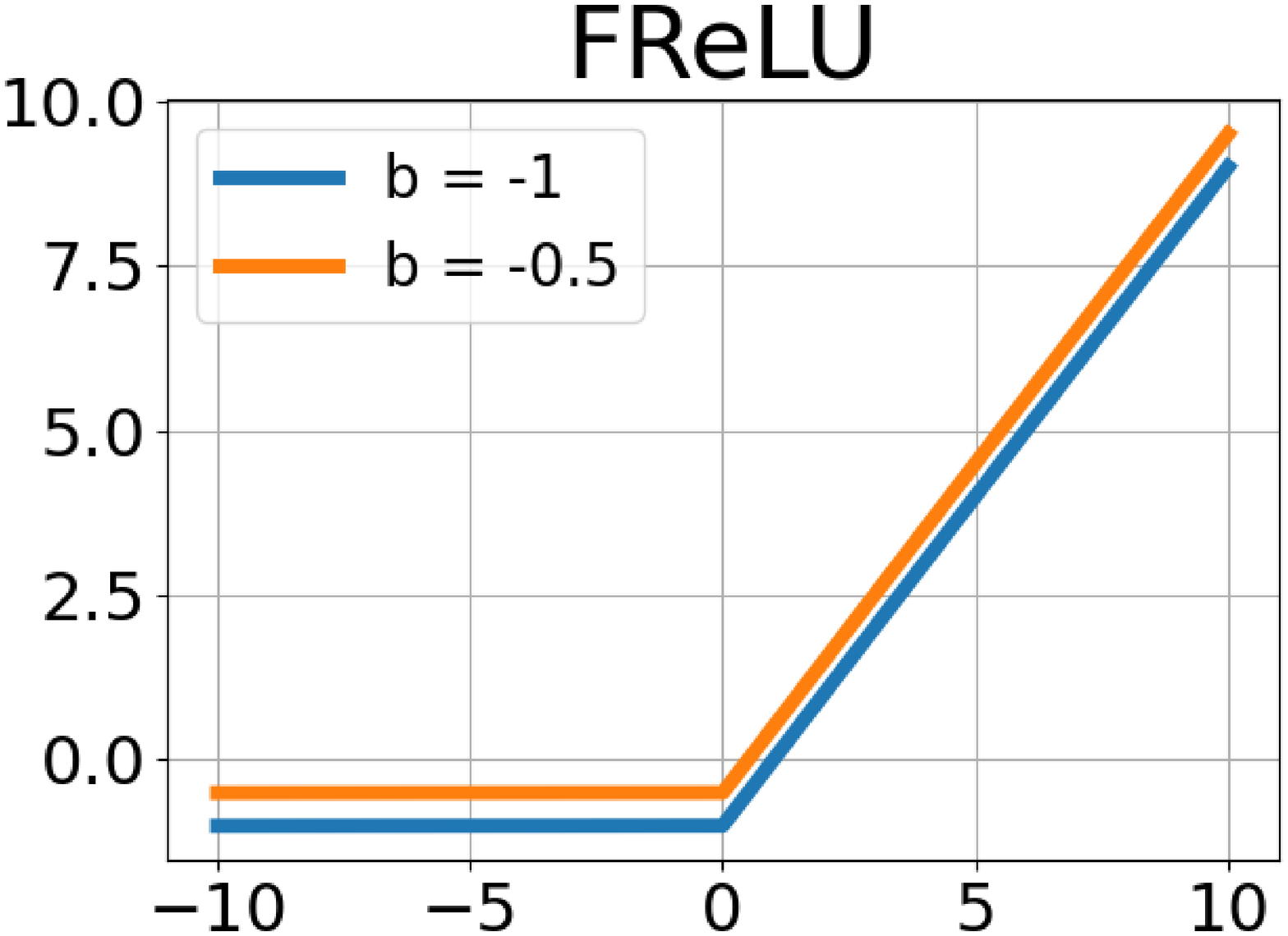}
    \includegraphics[width=1.5 in]{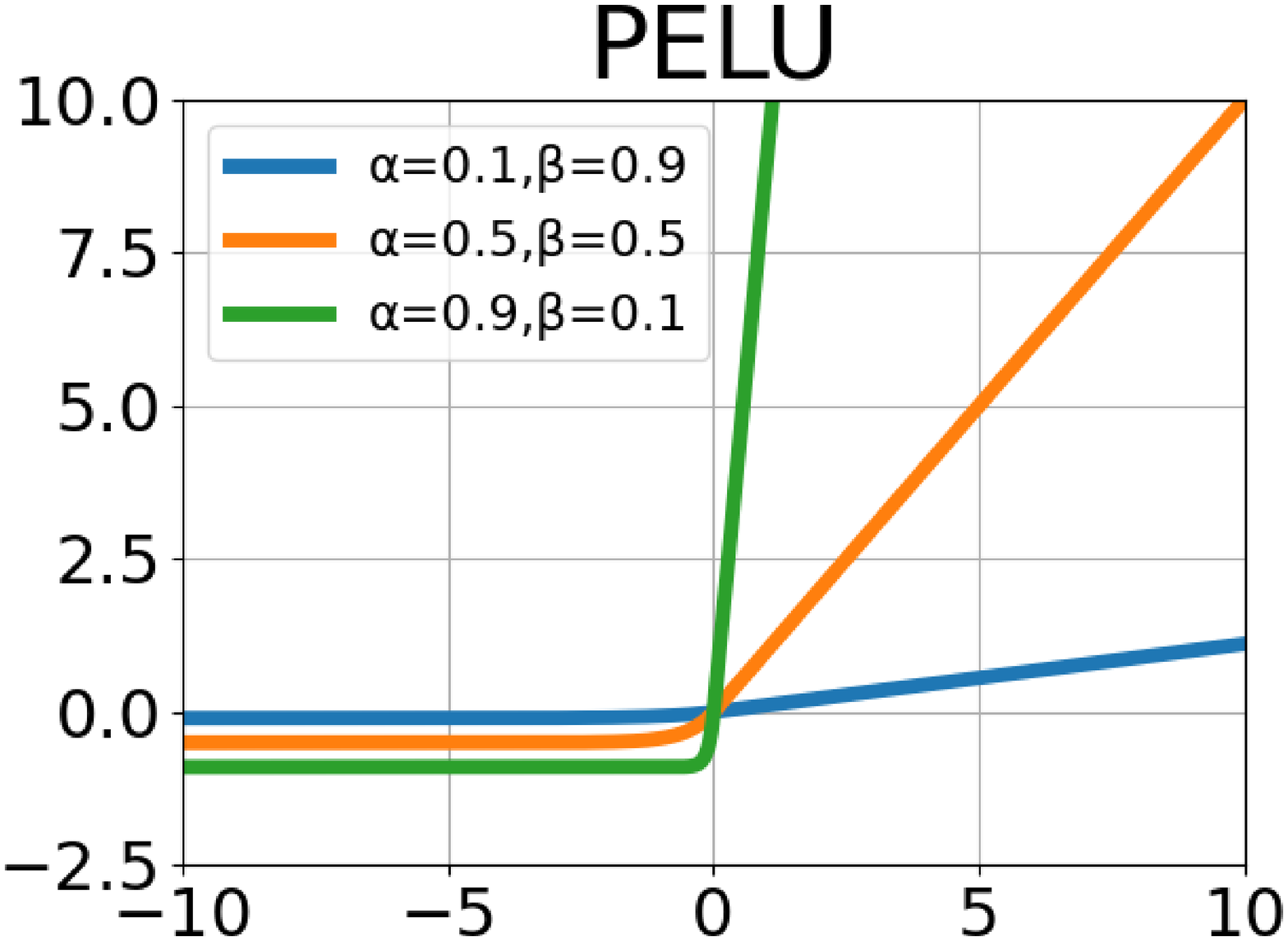}
	\caption{An illustration of various adaptive activation functions. The shapes of activation functions can be controlled and adjusted by some parameters, which are trained along with other parameters in the neural networks.}
	\label{fig:fig2}
\end{figure*}

Adaptive activation functions refer primarily to the functions that the parameters or shapes are trained and learned along with other parameters in neural networks (shown in Fig. \ref{fig:fig2}), thereby adaptively varying with training data. In other words, the main idea of this kind of functions is to search a good function shape using knowledge given by the training data. For example, PReLU \citep{he2015delving} replaces the fixed slope $\alpha$ of LReLU \citep{maas2013rectifier} with a trainable parameter $\alpha_i$ in the negative region. Whereas Swish \citep{ramachandran2017searching} is a recently proposed activation function with no upper bound, lower bound, smooth, and non-monotonic characteristic, and it can be loosely viewed as a bridging function between the linear function and the ReLU function. Other similar activation functions like FReLU \citep{qiu2018frelu} and PELU \citep{trottier2017parametric} have achieved performance improvements in some specific tasks.

Although the existing adaptive activation functions has shown to improve the network performances significantly, thanks to properties such as no saturation feature, flexibility and adaptivity, exploring the optimal and appropriate activation functions is still an open field of research, and there is still potential room for improvement in various scenarios, especially  for complex datasets and different models.

\section{Methodology}
The training of neural networks is essentially a non-convex optimization problem, in which the optimal weight parameters can be searched and found by using the back-propagation algorithm, so that the functional subspace will be explored and determined by the activation function. Adaptive activation functions refer to the functions that adapt themselves to the network inputs, therefore they can learn hyper-parameters to adapt the parameters of the affine transformation to a given input, and thereby increase the flexibility and the representation ability of network models.

In this work, we attempt to construct a new parameter learning method for each layer only by introducing a few parameters to the fixed activation functions, and the general form in the $i^{th}$ layer with activation functions can be defined as follows
\begin{equation}
f_A(a_i,b_i,c_i,d_i,z)=b_i f(a_i z+c_i)+d_i
\end{equation}
where $f(\cdot)$ represents a traditional activation function (fixed activation function). $a_i$, $b_i$, $c_i$ and $d_i$ are four learnable parameters in the $i^{th}$ layer, and they can adapt to the different tasks according to the complexity of input data so as to efficiently avoid falling into local minimums. $z$ denotes the weighted sum of inputs, including the bias term, defined as
\begin{equation}
z = w x+b
\end{equation}
$w$ and $b$ indicate weights and bias, respectively. $x$ is an input vector.

In practice, the proposed adaptive activation function is very simple, and it is composed of two embedded linear equations, namely: internal linear equation 
\begin{equation}f_{in}(a_i,c_i,z) = a_i z+c_i \end{equation}
and external linear equation
\begin{equation}f_{ex}(b_i,d_i) = b_i f_{in}+d_i \end{equation}
Therefore Equation (1) is rewritten as
\begin{equation}
f_A(a_i,b_i,c_i,d_i,z)=f_{ex}(b_i,d_i)=b_i f_{in}(a_i,c_i,z)+d_i
\end{equation}

In the following sections, the effectiveness and advantages of the proposed methodology will be verified by taking some common fixed activation functions as baselines, such as \textit{Sigmoid}, \textit{Tanh} and \textit{ReLU}, and the corresponding adaptive activation functions are named \textit{ASimoid}, \textit{ATanh} and \textit{AReLU}, respectively. According to Equation (5), these functions are respectively defined as  
\begin{equation}ASimoid: f_{Asimoid}=b_i Sigmoid(a_i z+c_i)+d_i\end{equation}
\begin{equation}ATanh: f_{Atanh}=b_i tanh(a_i z+c_i)+d_i\end{equation}
\begin{equation}AReLU: f_{Arelu}=maximum(a_i z+c_i, b_i z+d_i)\end{equation}
In ASigmoid and ATanh, $a_i$ and $c_i$ are respectively used to scale the inputs of Sigmoid and Tanh,while $b_i$ and $c_i$ scale the outputs,simultaneously.

Significantly, when $a_i=1$ and $b_i=c_i=d_i=0$, the negative part of the AReLU is replaced with a zero slope, while the slope of the positive part is fixed. In this case, AReLU is actually degenerated to a standard ReLU, given as 
\begin{equation}ReLU: f_{ReLU}=maximum(0, z)\end{equation}
Furthermore, when $b_i=1$ and $c_i=d_i=0$, the slope (i.e., the parameter $a_i$) of the negative part is adjustable, which means that the parameter can learn from data rather than be obtained by the pre-defined. Under these conditions, AReLU is evolved to PReLU when , given as  
\begin{equation}PReLU: f_{Prelu}=maximum(a z, z)\end{equation}
Therefore, \textit{AReLU} is a generalized form of the \textit{ReLU-based} versions, while \textit{ReLU} and \textit{PReLU} are the special cases of the proposed \textit{AReLU}.

Above, it can be clearly seen that our method only adds four parameters for each layer. For the entire network model, $4i$ parameters should be added. This parameter amount and calculation amount is negligible compared with the entire network model.

\begin{figure*}[!t]
	\centering
	\includegraphics[width=6in]{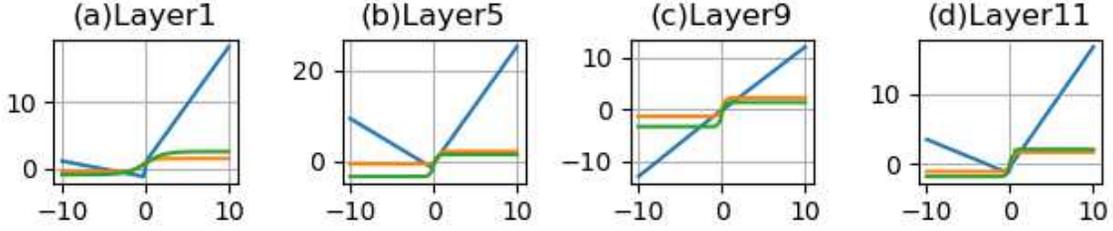}
	\caption{The shapes of ASigmoid (green), ATanh (orange) and AReLU (blue) at different layers during the training process on CIFAR100 based on VGG.}
	\label{fig:fig3}
\end{figure*}

\section{Analysis}
A convex loss function $L(\cdot)$ for the linear weighted combination of each activation function applied to an input is defined to find the optimal weights by adopting suitable optimization strategies based on the back-propagation algorithm. Thus, the training process of the network model is essentially an iterative optimization process for the weight parameters by minimizing the loss function $L(\cdot)$ in the functional subspace.
\subsection{Theoretical Analysis}
In order to facilitate analysis, we just take an activation function of one layer as an example without loss of generality. Suppose that a neural network with traditional activation functions is given as 
\begin{equation}y = f(z)\end{equation}
For the update process of weights, the partial derivative chain is defined as follows.
\begin{equation}\frac{\partial L}{\partial w}=\frac{\partial L}{\partial y} \frac{\partial y}{\partial z}\frac{\partial z}{\partial w}\end{equation}
\begin{equation}\frac{\partial y}{\partial z}=f'(z)\end{equation}
Meanwhile, the weight is updated as follows. 
\begin{equation}\widetilde{w}=w-\eta\frac{\partial L}{\partial\omega}\end{equation}
where $\eta$ is learning rate. Equation (16) and (17) are substituted into (18), we can obtain the weights update equation for a common activation function as follows.
\begin{equation}\widetilde{w}=w-\eta f'(z) \frac{\partial L}{\partial y}\frac{\partial z}{\partial w}\end{equation}

Considering that the proposed adaptive activation functions consist of two linear equations, for simplicity, we just consider the internal linear function and omit its intercept term in a certain layer, given as    
\begin{equation}\widetilde{y} = f(a z)\end{equation}
where $\widetilde{y}$ is the output of an adaptive activation function. The hyperparameter $a$ represents a generalization form for scale inputs in any layers, and it can be inferred to fine-tune the learning rate so as to speed up the update of weights, and the corresponding derivation is given as follows. 

For the output $\widetilde{y}$, the partial derivative is given as 
\begin{equation}\frac{\partial \widetilde{y}}{\partial z}=a f'(z)\end{equation}
With Equations (16), (18) and (21), the update process of the weight is given as 
\begin{equation}\widetilde{w}_1=w-\eta a f'(z) \frac{\partial L}{\partial \widetilde{y}}\frac{\partial z}{\partial w}\end{equation}
By comparison between Equations (19) and (22), the learning rate of the adaptive activation function can be written as:
\begin{equation}\widetilde{\eta}=\eta a\end{equation}
From Equation (23), we can adaptively adjust the learning rate $\widetilde{\eta}$ by using the hyperparameter $a$. Simultaneously, the optimization of the hyperparameter $a$ in neural networks is similar to the hyperparameter $w$, then the update process of $a$ is achieved by using the chain rule.

\begin{equation}\frac{\partial L}{\partial a}=\frac{\partial L}{\partial \widetilde{y}} \frac{\partial \widetilde{y}}{\partial a}\end{equation}
\begin{equation}\frac{\partial \widetilde{y}}{\partial a}=z f'(z) \end{equation}
With Equations (24) and (25)
\begin{equation}\widetilde{a}=a-\eta z f'(z) \frac{\partial L}{\partial \widetilde{y}}\end{equation}
With Equations (6) and (26)
\begin{equation}\widetilde{a}=a-\eta w x f'(z) \frac{\partial L}{\partial \widetilde{y}}\end{equation}
Therefore, the adaptive activation function is to achieve rapid convergence by the adapting learning rate, and this method is achieved by adjusting the weight $w$ and the parameter $a$ mutually to speed up learning in neural networks and lead to higher classification precision.

Besides, the internal linear equation has also its respective intercept, which contributes to tuning the parameters from another vertical direction during training process, thereby avoiding involving local extremum.

Similarly, the external linear equation has the same effects for accelerating the convergence and improving the performance. More importantly, the internal equation is embedded within the external, such case will enable the optimization toward a global optimum solution more efficiently from all directions. 

\subsection{Experimental Analysis}
Owing to the fact that each layer has its own independent activation function, the optimal hyperparameters of each layers can obtained by learning from the respective complexity of input data, and their values will vary with the input data characteristics. Therefore, the obtained weights will be optimal, and the corresponding activation functions are different with different layers. Fig. \ref{fig:fig3} shows the visualization of different layers during the training process on CIFAR100 based on VGG, and different shapes of ASigmoid, ATanh and AReLU at different layers indicate that these functions can learn the optimal hyperparameters from the inputs of the respective layers, which would lead to the enhancement and improvement of the fitting capability and the accuracy of the networks.

\begin{figure*}[!t]
	\centering
	\includegraphics[width=6.3in]{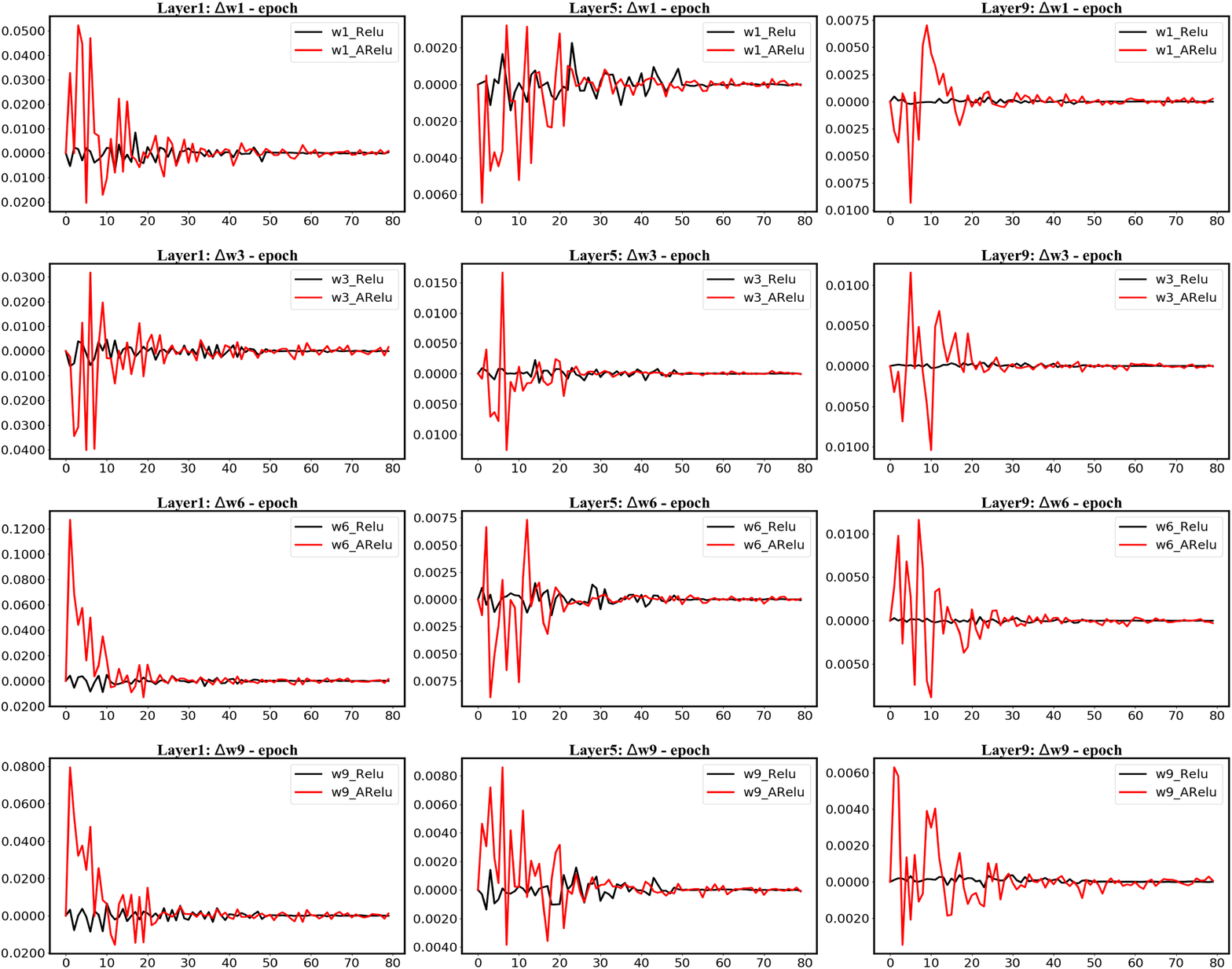}
	\caption{The change curves of the weight increments $\Delta w$ between two successive epochs in different layers during the training process by using AReLU and ReLU on CIFAR100 based on VGGNet. Column 1,2 and 3 represent the $1^{th}$, $5^{th}$ and $9^{th}$ layer, respectively. Rows 1-4 illustrate four various weights by using AReLU and ReLU.}
	\label{fig:fig4}
\end{figure*}

Moreover, compared with the traditional adaptive activation functions, two embedded linear equations with intercepts can accelerate the weights adjustment. For further verification, the change curves of the weight increments $\Delta w$ between two successive epochs in various layers are visualized along with the training process (shown in Fig. \ref{fig:fig4}). The results clearly show the amplitudes of the increments $\Delta w$ by using the proposed AReLU are much larger than those of the traditional ReLU in the early training stages, then the increments of the two methods converge, which means the proposed methods can provide faster weight updates than the traditional methods. Consequently the proposed methods can improve greatly the convergence speed and reduce the computational burden. Meanwhile, the large amplitudes of the increments can also help to avoid falling into a local optimum when training artificial neural networks with gradient-based learning methods and backpropagation.

\section{Experiments}
In this section, a series of experiments are implemented to verify and evaluate the effectiveness of the proposed methodology based on the three baseline activation functions such as Sigmoid, Tanh and ReLU. Considering the fact that ReLU is the most common activation function used in neural networks, there exist many derivatives of ReLU, and some typical derivatives like LReLU and PReLU are selected for comparison to highlight the effectiveness of the parameterization method in the activation function. Whereas swish, as an outstanding activation function, is used to demonstrate the state-of-the-art performance of the proposed adaptive activation function. Firstly, many comparison experiments between the proposed functions and its corresponding baseline functions have been conducted by using Stochastic gradient descent (SGD) \citep{cramer1946mathematical} on the datasets of CIFAR10 and CIFAR100 based on different network models, such as AlexNet \citep{krizhevsky2012imagenet}, VGG \citep{simonyan2014very}, GoogleNet \citep{szegedy2015going}, ResNet \citep{he2016deep} and DenseNet \citep{huang2017densely}. Then, some experiments are implemented to further verify the validity and suitability in various optimization strategies, such as SGD, Momentum \citep{qian1999momentum}, AdaGrad \citep{duchi2011adaptive}, AdaDelta \citep{zeiler2012adadelta} and ADAM \citep{kingma2014adam}. Finally, a series of comparison experiments are conducted to further verify the effectiveness, suitability and generalization ability on other more complicated datasets like miniImageNet\citep{Oriol2016}, PASCAL VOC \citep{VOC2012} and COCO \citep{COCO2014}.

\subsection{Experimental setup}
\begin{figure*}[!t]
	\centering
	\includegraphics[width=6in]{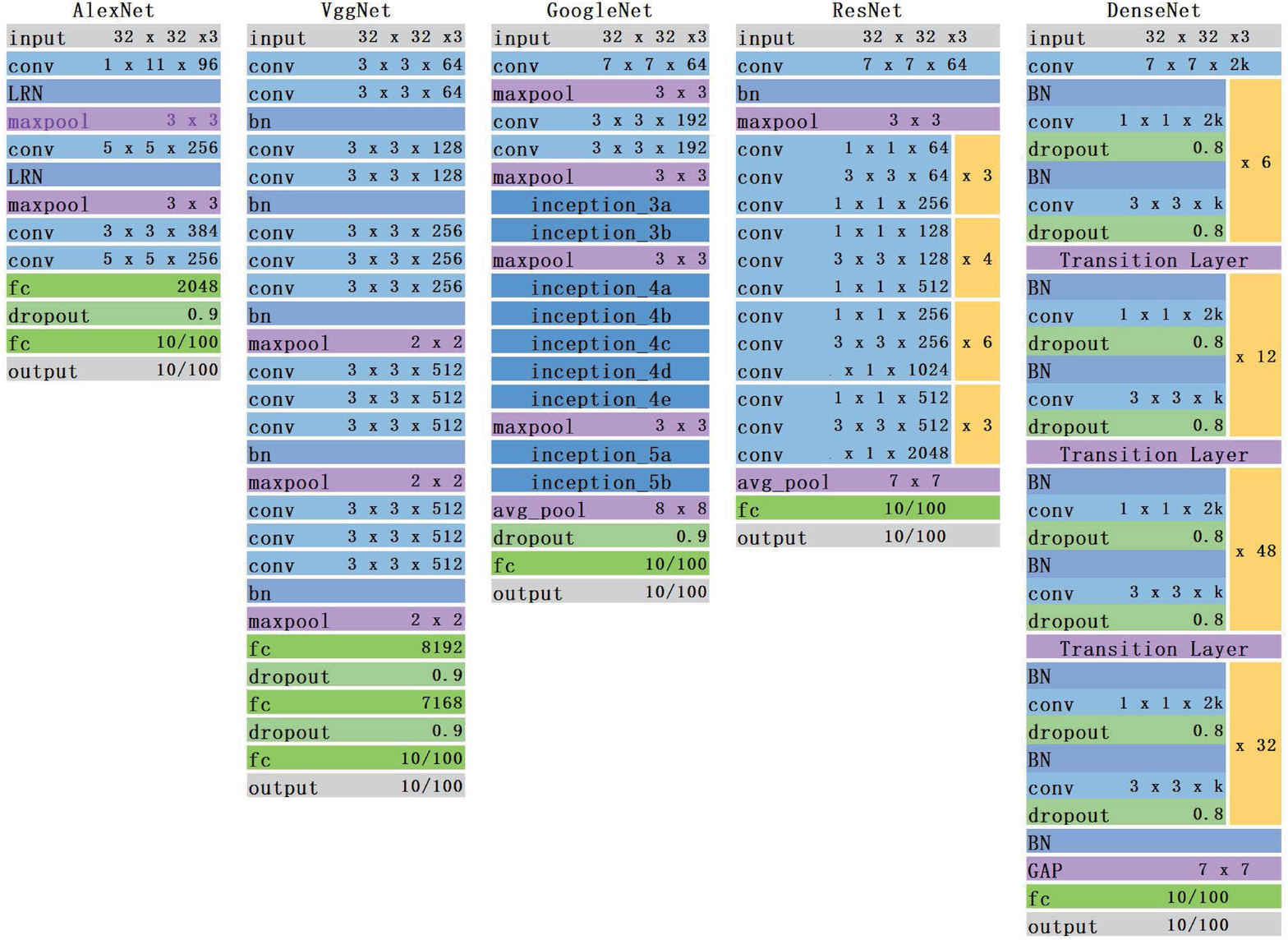}
	\caption{The network architectures of AlexNet, VGGNet, GoogleNet, ResNet and DenseNet on CIFAR10 and CIFAR100.}
	\label{fig:fig5}
\end{figure*}

We test the proposed adaptive activation functions on CIFAR10 and CIFAR100 based on AlexNet, VGGNet, GoogleNet, ResNet and DenseNet. The detailed experimental setup is illustrated in Fig. \ref{fig:fig5}. 

Note that Dense block consists of $\{\stackrel{bn} \longrightarrow conv(1, 1, 2k)  \stackrel{drop0.8, bn}\longrightarrow conv(3, 3, 2k)\stackrel{drop0.2}\longrightarrow \}\times 6$, $\times 12$, $\times 48$, $\times 32$, respectively, and the transition layer is shown in Fig. \ref{fig:fig3}, which corresponds to the sequence $conv(1, 1) - avg\_poo(2, 2).$  Moreover, the growth rate is k=24 for all.

\subsection{CIFAR10}
In these experiments, the proposed adaptive activation functions (10)$\sim$(12) are applied on CIFAR10 dataset based on the models (shown in Fig. \ref{fig:fig5}), respectively. All trainings are implemented for no less than 80 epochs with a 64-batch size and without data augmentation by using SGD with fixed learning rate schedule of 0.001, 0.0001 and 0.00001 with the training process, respectively. 

\begin{figure*}[!t]
	\centering
	\includegraphics[scale=0.24]{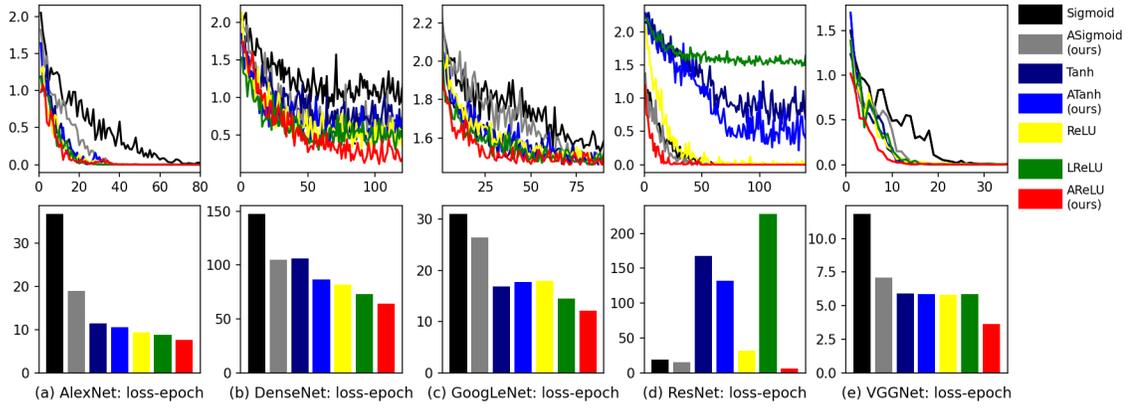}
	\caption{The top row represents the loss-epoch convergence curves for various activation functions on CIFAR10 by using SGD optimizer on various models during the training process, such as AlexNet, DenseNet, VGGNet, ResNet and DenseNet. Whereas the bottom row illustrates the area enclosed by the corresponding convergence curves in top row,  and the smaller area means the faster convergence speed.}
	\label{fig:fig6}
\end{figure*}

Fig. \ref{fig:fig6} shows the convergence curves (top row) and the area enclosed by the convergence curves (bottom row) during the training process. It is obvious that the smaller the area, the faster the convergence speed, and it is also clearly shown from the results that the proposed activation functions can surpass the corresponding baseline functions and LReLU for different network models in terms of convergence speed. Thereinto, AReLU can obtain the fastest convergence speed among these activation functions, especially on DenseNet, ResNet and VGG16, it can converge much faster than other activation functions. Tables~\ref{tab:table1}  and Table~\ref{tab:table2}\footnote{Indicates that the activation function does not converge in this mode.} show the quantified results in precision, and the proposed methodology has an overall advantage. Table~\ref{tab:table1} illustrates the comparison results between the proposed methods and their respectively corresponding baselines. From the results, the proposed methodology can effectively applied to the classic fixed activation functions, and can surpass the corresponding baseline functions on most network models. For instance, ASigmoid can surpass the corresponding baseline function Sigmoid on all models, while ATanh can also obtain better precision than the corresponding Tanh on the models of VGGNet, ResNet and DenseNet, and the obtained precision of ATanh in AlexNet and GoogleNet are only slightly lower than those of the corresponding Tanh. Table~\ref{tab:table2} shows the comparison results between AReLU and other adaptive activation functions. AReLU, except PReLU on the AlexNet and VGGNet, can overall obtain higher precision than other adaptive functions on various models. Note that some traditional adaptive functions like PELU and FReLU are not suitable for some network models owing to the lack of convergence during the training process. While the proposed methodology can apply to various deep learning models and has better generalization performance than other traditional methodologies.

\begin{table*}[!t]
\renewcommand{\arraystretch}{1.3}
\caption{The classification precision of various fixed activation functions for different models on CIFAR10.}
\centering
\resizebox{\textwidth}{!}{
\begin{tabular}{lccccc}
\toprule
Methods   & AlexNet & VGGNet & GoogleNet & ResNet & DenseNet \\
\midrule
Sigmoid  & 0.8141$\pm$1.52e-3 & 0.8680$\pm$1.17e-3 & 0.8027$\pm$1.13e-3 & 0.8079$\pm$1.25e-3 & 0.7611$\pm$3.86e-3 \\
ASigmoid(ours)  & \myfontb{0.8469$\pm$1.68e-3} & \myfontb{0.8819$\pm$2.28e-4} & \myfontb{0.8517$\pm$4.07e-4} & \myfontb{0.8087$\pm$1.36e-3} & \myfontb{0.9076$\pm$8.16e-4} \\
\hline
Tanh  & \myfontb{0.8043$\pm$9.68e-4} & 0.8841$\pm$7.72e-4 & \myfontb{0.8433$\pm$6.06e-4} & 0.5784$\pm$9.75e-4 & 0.8966$\pm$1.99e-5 \\
ATanh(ours) & 0.7900$\pm$5.31e-4 & \myfontb{0.8904$\pm$3.05e-4} & 0.8342$\pm$7.99e-4 & \myfontb{0.6014$\pm$8.11e-4} & \myfontb{0.9163$\pm$1.87e-4} \\
\hline
ReLU  & \myfontb{0.8351$\pm$3.95e-4} & 0.8800$\pm$1.01e-4 & 0.8733$\pm$4.36e-5 & 0.6641$\pm$2.59e-3 & 0.9242$\pm$1.10e-4 \\
LReLU\citep{maas2013rectifier}  & 0.8255$\pm$1.66e-3 & \myfontb{0.9253$\pm$4.68e-5} & 0.8742$\pm$9.31e-4 & 0.6698$\pm$1.27e-3 & 0.9289$\pm$2.37e-5 \\
AReLU(ours) & 0.8331$\pm$2.48e-4 & 0.8328$\pm$1.26e-3 &  \myfontb{0.8773$\pm$1.83e-3} &  \myfontb{0.9230$\pm$5.51e-4} &  \myfontb{0.9538$\pm$5.58e-4} \\
\bottomrule
\end{tabular}
}
\label{tab:table1}
\end{table*}

\begin{table*}[!t]
\renewcommand{\arraystretch}{1.3}
\caption{The classification precision of various adaptive adaptive activation functions for different models on CIFAR10. The top results are highlighted in black bold and the second-best results in
blue.}
\centering
\begin{threeparttable}
\resizebox{\textwidth}{!}{
\begin{tabular}{lccccc}
\toprule
Methods   & AlexNet & VGGNet & GoogleNet & ResNet & DenseNet \\
\midrule
PReLU\citep{he2015delving} & \myfontb{0.8558$\pm$4.61e-4} & \myfontb{0.9344$\pm$5.89e-5} & 0.8551$\pm$1.79e-4 & 0.6522$\pm$3.20e-4 & 0.9231$\pm$1.40e-4 \\
Swish\citep{ramachandran2017searching}  & 0.7557$\pm$2.20e-3 &\textcolor{blue}{0.9171$\pm7.98$e-4} &\textcolor{blue}{0.8710$\pm$3.25e-3}& 0.6446$\pm$3.39e-4 & \textcolor{blue}{0.9276$\pm$1.87e-4} \\
PELU\citep{trottier2017parametric} & $-$\tnote{1} &   $-$   & 0.8128$\pm$4.45e-4 & 0.7891$\pm$3.91e-4 & $-$      \\
FReLU\citep{qiu2018frelu} & $-$  & 0.8558$\pm$2.44e-4 & 0.8694$\pm$5.21e-3 &\textcolor{blue}{0.8726$\pm$3.71e-4} & 0.8214$\pm$4.11e-4 \\
AReLU(ours) & \textcolor{blue}{0.8331$\pm$2.48e-4}& 0.8328$\pm$1.26e-3 &  \myfontb{0.8773$\pm$1.83e-3} &  \myfontb{0.9230$\pm$5.51e-4} &  \myfontb{0.9538$\pm$5.58e-4} \\
\bottomrule
\end{tabular}
}
\end{threeparttable}
\label{tab:table2}
\end{table*}

\subsection{CIFAR100}
To further verify the validity and applicability, CIFAR-100 is selected to training based on several typical network models, such as AlexNet, DenseNet (shown in Fig. \ref{fig:fig5}) and VGG-v. The Dataset is trained for 150 epochs with a 250-batch size and a fixed learning rate of 0.001, 0.0001 and 0.00001 with the training process, respectively. Note that VGG-v is an extended version of the VGGNet.

Table~\ref{tab:table3} shows the classification comparison results between the proposed methods and their corresponding baselines on AlexNet, VGG-v and DenseNet, respectively. Except AReLU in AlexNet, the proposed methods can surpass the respective corresponding baseline functions. Table~\ref{tab:table4}\footnote{Indicates that the activation function does not converge in this method.} illustrates the comparison results between AReLU and other adaptive functions. From the results, AReLU can achieve the best precision performance on the three network models. Similarly, PELU and FReLU cannot converge to desired loss values.

\begin{table*}[!t]
\renewcommand{\arraystretch}{1.3}
\caption{The classification precision of various fixed activation functions for different models on CIFAR100.}
\centering
\begin{tabular}{lcccc}
\toprule
Methods & AlexNet & VGG-v & DenseNet \\
\midrule
Sigmoid & 0.4662$\pm$2.49e-3 & 0.5545$\pm$5.89e-4 & 0.2561$\pm$1.23e-3 \\
ASigmoid(ours) & \myfontb{0.5312$\pm$8.66e-4} & \myfontb{0.6578$\pm$7.56e-4} & \myfontb{0.6103$\pm$3.15e-3} \\
\hline
Tanh & 0.5058$\pm$6.06e-4 & 0.6007$\pm$6.90e-4 & 0.5960$\pm$2.53e-3   \\
ATanh(ours) & \myfontb{0.5236$\pm$1.98e-4} & \myfontb{0.6166$\pm$2.41e-4} & \myfontb{0.6734$\pm$3.98e-4}  \\
\hline
ReLU &\myfontb{0.5701}$\pm$1.18e-4 & 0.6972$\pm$4.99e-4 & 0.5616$\pm$1.09e-3  \\
LReLU\citep{maas2013rectifier} & 0.5500$\pm$1.09e-3 & 0.6991$\pm$5.53e-4 & 0.6914$\pm$5.12e-4 \\
AReLU(ours) & 0.5647$\pm$1.71e-4 & \myfontb{0.7005$\pm$1.04e-3} & \myfontb{0.7081$\pm$1.64e-3}   \\
\bottomrule
\end{tabular}
\label{tab:table3}
\end{table*}

\begin{table*}[!t]
\renewcommand{\arraystretch}{1.3}
\caption{The classification precision of various adaptive activation functions for different models on CIFAR100.}
\centering
\begin{threeparttable}
\begin{tabular}{lcccc}
\toprule
Methods & AlexNet & VGG-v & DenseNet \\
\midrule
PReLU\citep{he2015delving} &0.5325$\pm$5.01e-4 & 0.6838$\pm$3.84e-5 & 0.5781$\pm$8.87e-3 \\
Swish\citep{ramachandran2017searching} &0.5519$\pm$1.16e-3 & 0.6729$\pm$8.32e-5 & 0.7079$\pm$3.01e-3   \\
PELU\citep{trottier2017parametric} & $-$\tnote{1} & $-$ &  $-$ \\
FReLU\citep{qiu2018frelu} & $-$ & 0.1534$\pm2.17e-4$ & 0.1325$\pm$2.16e-4  \\
AReLU(ours) & \myfontb{0.5647$\pm$1.71e-4} & \myfontb{0.7005$\pm$1.04e-3} & \myfontb{0.7081$\pm$1.64e-3}   \\
\bottomrule
\end{tabular}
\end{threeparttable}
\label{tab:table4}
\end{table*}

\subsection{Validity and practicability in various optimization strategies}
Gradient descent algorithms are often used as a black-box optimizer in neural networks, and different optimization strategies have great influence on the performance of activation functions in practice. Therefore, to further verify the validity and practicability in various optimization strategies, the best method AReLU is selected as activation function based on GoogleNet and ResNet, and a series of comparison experiments are achieved on CIFAR10 among various optimizers, such as SGD, Momentum, AdaGrad, AdaDelta and ADAM. 

\begin{figure*}[!t]
	\centering
    \includegraphics[width=6in]{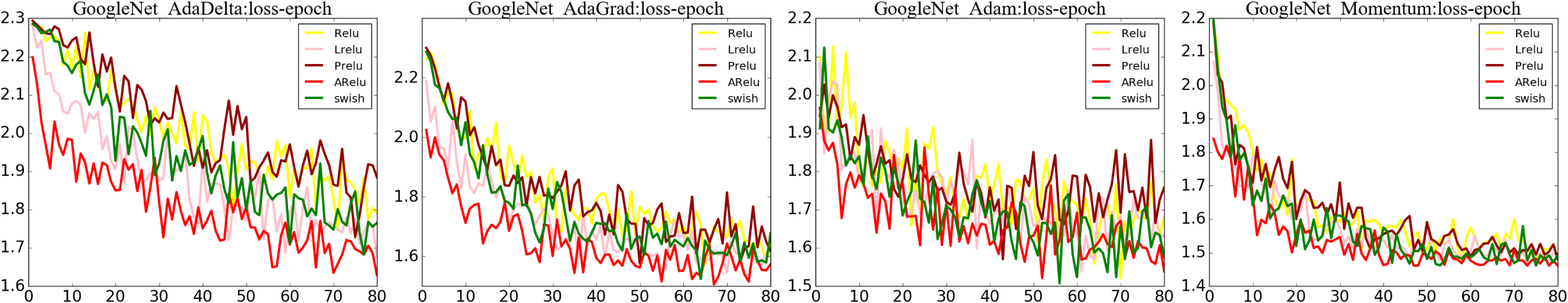}
    \includegraphics[width=6in]{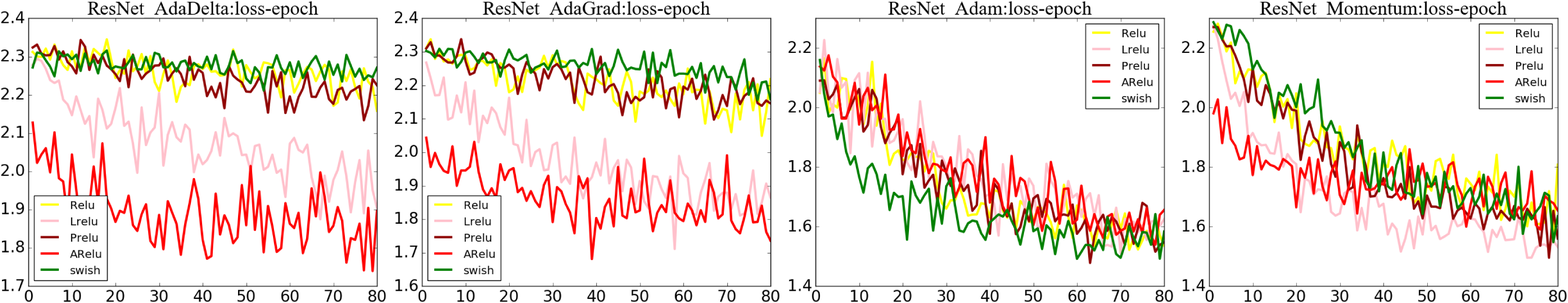}
	\caption{The convergence curves by using various optimization strategies on GoogleNet (top row) and  ResNet (bottom row).}
	\label{fig:fig7}
\end{figure*}

Fig. \ref{fig:fig7} shows the convergence curves by using various activation functions with various optimization strategies on GoogleNet and ResNet, respectively. The results show that AReLU converges faster than all other activation functions on the GoogleNet model. Whereas on the ResNet model, it is also obvious for AReLU to have an overall convergence advantage, especially compared with AdaGrad and AdaDelta. These results indicate that the proposed AReLU can accelerate convergence, thereby reducing the training cost.  

\begin{figure*}[!t]
	\centering
	\includegraphics[width=6in]{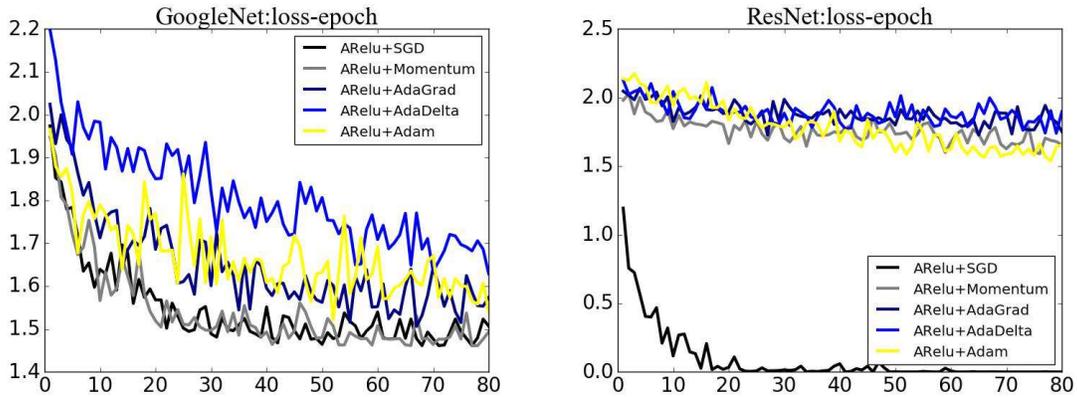}
	\caption{The convergence curves of AReLU by using various optimization strategies on GoogleNet and ResNet.}
	\label{fig:fig8}
\end{figure*}

Table~\ref{tab:table5} further reveals that the proposed AReLU can achieve better overall performance than other activation functions based on different optimization strategies and different network models. Except ReLU with a Momentum optimizer on GoogleNet and Swish with an ADAM optimizer on ResNet, the proposed AReLU surpasses other activation functions with various optimization strategies on both the network models, and the obtained precision performance is far better than other methods. While AReLU is only slightly worse than ReLU with a Momentum optimizer on GoogleNet and Swish with an ADAM optimizer on ResNet, respectively. Significantly, AReLU with SGD can generally achieve the best precision performance among these activation functions with various optimization strategies on both models. Fig. \ref{fig:fig8} illustrates the convergence of the proposed AReLU with various optimizers, and the results show that SGD has faster convergence speed than all other optimizers on both models, especially on ResNet.

The above results shows the proposed adaptive activation functions have faster convergence speed and higher precision than traditional activation functions. And it suggests the proposed methodology can avoid local minimums and accelerate convergence, thereby increasing the precision, reducing the training cost and improving the generalization performance.

\begin{table*}[!t]
\renewcommand{\arraystretch}{1.3}
\caption{Classification precision comparisons between various activation functions by using different optimization strategies and models on CIFAR10. The top results are highlighted in black bold and the second-best results in blue.}
\centering
\resizebox{\textwidth}{!}{
\begin{tabular}{lcccccc}
\toprule
Models & Methods & SGD & Momentum & AdaGrad & AdaDelta & ADAM \\
\midrule
           &ReLU  & 0.8733$\pm$4.36e-5          &\myfontb{0.8951$\pm$7.30e-5}   &0.6065$\pm$1.53e-4           &0.5686$\pm$3.31e-4          &\textcolor{blue}{0.8797$\pm$1.08e-3}  \\
          &LReLU\citep{maas2013rectifier} &\textcolor{blue}{0.8742$\pm$9.31e-4} & 0.8498$\pm$4.20e-4 &0.6623$\pm$3.22e-3 &\textcolor{blue}{0.6521$\pm$1.92e-3} &0.8654$\pm$8.78e-5\\
GoogleNet &PReLU\citep{he2015delving} &0.8551$\pm$1.79e-4  &0.8138$\pm$2.14e-4 &0.6278$\pm$8.44e-6 &0.5367$\pm$1.45e-3 &0.8177$\pm$2.34e-6\\
          &Swish\citep{ramachandran2017searching} &0.8710$\pm$3.25e-3  &\textcolor{blue}{0.8772$\pm$1.39e-3} &\textcolor{blue}{0.6747$\pm$1.58e-3}  &0.6139$\pm$2.66e-4  &0.8743$\pm$1.92e-3  \\
		 &AReLU(ours) &\myfontb{0.8773$\pm$1.83e-3}  & 0.8651$\pm$8.44e-5  &\myfontb{0.7408$\pm$6.38e-3}  &\myfontb{0.7221$\pm$2.57e-3} &\myfontb{0.8910$\pm$3.37e-4} \\
\hline
          &ReLU  & 0.6641$\pm$2.59e-3         &0.6274$\pm$1.99e-3          & 0.4253$\pm$7.38e-4        &0.3553$\pm$1.57e-3        &0.8519$\pm$6.20e-4\\
          &LReLU\citep{maas2013rectifier} &\textcolor{blue}{0.6698$\pm$1.27e-3} &\textcolor{blue}{0.6635$\pm$1.40e-3} &\textcolor{blue}{0.5162$\pm$1.61e-4} &\textcolor{blue}{0.4046$\pm$3.87e-3} &0.8065$\pm$1.07e-3\\
ResNet  & PReLU\citep{he2015delving} & 0.6522$\pm$3.20e-4&0.6325$\pm$2.11e-4 & 0.3493$\pm$2.07e-3 & 0.2846$\pm$4.28e-3 &0.7889$\pm$8.08e-4\\
		   &Swish\citep{ramachandran2017searching} & 0.6446$\pm$3.39e-4        &0.6099$\pm$7.02e-5          & 0.4032$\pm$3.41e-3        &0.3112$\pm$3.31e-3         &\myfontb{0.8909$\pm$6.84e-5} \\
		   &AReLU(ours) & \myfontb{0.9230$\pm$5.51e-4} &\myfontb{0.7862$\pm$1.38e-3} & \myfontb{0.7341$\pm$2.92e-3}&\myfontb{0.7155$\pm$5.56e-4}&\textcolor{blue}{0.8805$\pm$2.31e-4} \\
\bottomrule
\end{tabular}
}
\label{tab:table5}
\end{table*}

\subsection{More complicated datasets}
Two more complicated datasets miniImageNet\citep{Oriol2016} and PASCAL VOC\citep{VOC2012} are used to further test the validity of the proposed methodology based on ResNet50. Table~\ref{tab:table6}\footnote{Indicates that the activation function does not converge in this method.} shows the performance comparisons of classification precision between AReLU and other adaptive functions.  The results indicate that AReLU and PReLU can obtain the best classification precision in PASCAL VOC and miniImageNet, respectively.  Overall, their performances are nearly equal in the two datasets. Note that PELU and FReLU cannot converge, meaning that the two datasets are very complicated and challenging.  \par  

The above all experiments are conducted for various models, datasets and methods based on classification tasks, and it means that the proposed adaptive activation functions can overall obtain the best classification performance. To test the validity and practicability in other deep learning tasks, PASCAL VOC is used for the object detection tasks by respectively adopting Faster RCNN\citep{RenHGS15} and YOLOv2\citep{redmon2016yolo9000} based on the proposed AReLU. Besides,  a more complicated detection dataset COCO\citep{COCO2014} is selected to further verify the effectiveness by employing FCOS\citep{tian2021fcos}. Table~\ref{tab:table7}\footnote{Activation function does not converge in this model.} shows performance comparisons of detection precision among various adaptive functions. From the results, the proposed AReLU can nearly achieve the best detection performance including AP50, AP75 and mAP among various adaptive functions based on various methods and datasets. It means that our method has a validity and practicability. \par

From the results of a series of comparison experiments, the proposed methods can achieve better performance in various scenarios, such as datasets, network models, optimization methods and deep learning tasks. The most significant reason is that our methodology has an internal and external bilinear structure, and the internal function is embedded within the external, such case will enable the optimization toward a global optimum solution more efficiently from all directions, thereby accelerating the convergence and improving the performance. More importantly, the proposed methodology only adds a small number of parameters (i.e., four parameters for each layer) , and the number of parameters is negligible compared with millions of parameters in entire network model, which means the amount of network computing and the risk of over-fitting is only increased inconsiderably.

\begin{table*}[!t]
\renewcommand{\arraystretch}{1.3}
\caption{The classification precision of various activation functions on miniImageNet and PASCAL VOC based on ResNet50. The top results are highlighted in black bold and the second-best results in blue.}
\centering
\begin{threeparttable}
\begin{tabular}{lcc}
\hline
Methods                                                   & miniImageNet                    & PASCAL VOC  \\
\midrule
ReLU                                                       &  0.7562$\pm$4.60e-3         & 0.5933$\pm$5.70e-3    \\
PReLU\citep{he2015delving}                      &\myfontb{0.7813$\pm$1.20e-3}          & \textcolor{blue}{0.6066$\pm$2.00e-4}    \\
Swish\citep{ramachandran2017searching}  &0.7134$\pm$2.50e-3             & 0.5154$\pm$2.30e-3     \\
PELU\citep{trottier2017parametric}           & $-$\tnote{1}                       & $-$                                \\
FReLU\citep{qiu2018frelu}                        & $-$                                    & $-$                                  \\
AReLU(ours)                                           & \textcolor{blue}{0.7751$\pm$1.40e-3}          & \myfontb{0.6092$\pm$3.02e-3}    \\
\hline
\end{tabular}
\end{threeparttable}
\label{tab:table6}
\end{table*}

\begin{table*}[!t]
\renewcommand{\arraystretch}{1.3}
\caption{The detection precision of various activation functions for different methods on various datasets. The top results are highlighted in black bold and the second-best results in blue.}
\centering
\begin{threeparttable}
\resizebox{\textwidth}{!}{
\begin{tabular}{lccc|ccc|ccc}
\hline
\multicolumn{1}{c}{\multirow{2}{1.2cm}{Methods}} & \multicolumn{3}{c|}{Faster-RCNN (PASCAL VOC)} & \multicolumn{3}{c|}{YOLOv2 (PASCAL VOC)} & \multicolumn{3}{c}{FCOS (COCO)}\\
\cline{2-10}
\multicolumn{1}{c}{}&\multicolumn{1}{c}{AP50} &\multicolumn{1}{c}{AP75} & \multicolumn{1}{c|}{mAP} & \multicolumn{1}{c}{AP50} & \multicolumn{1}{c}{AP75} &\multicolumn{1}{c|}{mAP} & \multicolumn{1}{c}{AP50} & \multicolumn{1}{c}{AP75} & \multicolumn{1}{c}{mAP}\\
\midrule
ReLU &\textcolor{blue}{0.8501} & 0.6266 & 0.7661 &0.5924 & 0.1459 & 0.3970 &0.4974 & 0.3426 & 0.3232\\
PReLU\citep{he2015delving} & $-$\tnote{1} & $-$  & $-$  & 0.1777 & 0.0176 & 0.0931 & 0.4943 & 0.3399 & 0.3214\\ 
Swish\citep{ramachandran2017searching} & 0.7111 & 0.4981 & 0.6658  &\textcolor{blue}{0.6250} &\textcolor{blue}{0.2432} &\textcolor{blue}{0.4621} & 0.4923 &0.3418 & 0.3220 \\
PELU\citep{trottier2017parametric} & $-$ & $-$  & $-$  & $-$ & $-$ & $-$ & $-$ & $-$ & $-$\\ 
FReLU\citep{qiu2018frelu} & 0.8486 & \myfontb{0.6273} &\textcolor{blue}{0.7669} & 0.2604 & 0.0627 & 0.1670 &\myfontb{0.5030} &\textcolor{blue}{0.3473} &\textcolor{blue}{0.3274} \\
AReLU(ours) &\myfontb{0.8530} &\textcolor{blue}{0.6271} & \myfontb{0.7675} & \myfontb{0.6384} & \myfontb{0.2459} & \myfontb{0.4687} & \textcolor{blue}{0.5009} & \myfontb{0.3490} & \myfontb{0.3279}\\
\hline
\end{tabular}
}
\end{threeparttable}
\label{tab:table7}
\end{table*}

\section{Conclusions}
In this work, a novel methodology is proposed to adaptively customize activation functions for various layers, and it will contribute to avoiding local minimums and accelerating convergence, thereby increasing the precision, reducing the training cost and improving the generalization performance. In this methodology, a small number of parameters are introduced to the traditional activation functions such as Sigmoid, Tanh and ReLU, and some theoretical and experimental analysis for accelerating the convergence and improving the performance is presented. To verify the effectiveness of the proposed methodology, a series of experiments are implemented on CIFAR10, CIFAR100, miniImageNet, PASCAL VOC and COCO by employing various network models such as VGGNet, GoogleNet, ResNet, DenseNet, various optimization strategies such as SGD, Momentum, AdaGrad, AdaDelta and ADAM, and various task like classification and detection tasks. The results show that the proposed methodology is very simple but with significant performance in convergence speed, precision and generalization, and it can surpass other popular methods like ReLU and swish in almost all experiments in terms of overall performance.

 \section{Acknowledgments}
The authors would like to express their appreciation to the referees for their helpful comments and suggestions.
This work was supported in part by Zhejiang Provincial Natural Science Foundation of China (Grant No. LGF20H180002 and GF22F037921), and in part by National Natural Science Foundation of China (Grant No. 61802347, 61801428 and 61972354), and the National Key Research and Development Program of China (Grant No. 2018YFB1305202).

\bibliographystyle{unsrtnat}

\bibliography{my_paper}

\end{document}